\documentclass[10pt,twocolumn,letterpaper]{article}

\usepackage{cvpr}
\usepackage{times}
\usepackage{epsfig}
\usepackage{graphicx}
\usepackage{amsmath}
\usepackage{amssymb}
\usepackage{enumerate}
\usepackage{enumitem}
\usepackage{color}
\usepackage{comment}

\usepackage[pagebackref=true,breaklinks=true,letterpaper=true,colorlinks,bookmarks=false]{hyperref}

\cvprfinalcopy 

\def\cvprPaperID{3606} 

\ifcvprfinal\fi
\begin{document}

\newcommand{\gscom}[1]{\textcolor{magenta}{\small {#1}}}

\newcommand{\changeurlcolor}[1]{\hypersetup{urlcolor=#1}}  

\title{Deep Back-Projection Networks For Super-Resolution}

\author{Muhammad Haris$^1$, Greg Shakhnarovich$^2$, and Norimichi Ukita$^1,$\\
$^1$Toyota Technological Institute, Japan $^2$Toyota Technological Institute at Chicago, United States\\
{\tt\small \{mharis, ukita\}@toyota-ti.ac.jp, greg@ttic.edu}
}

\maketitle

\begin{abstract}
The feed-forward architectures of recently proposed deep
super-resolution networks learn representations of low-resolution
inputs, and the non-linear mapping from those to high-resolution
output. However, this approach does not fully address the mutual
dependencies of low- and high-resolution images. We propose Deep
Back-Projection Networks (\changeurlcolor{blue}\href{http://www.toyota-ti.ac.jp/Lab/Denshi/iim/members/muhammad.haris/projects/DBPN.html}{DBPN}), that exploit iterative up- and down-sampling
layers, providing an error feedback mechanism for projection errors at
each stage. We construct mutually-connected up- and down-sampling
stages each of which represents different types of image degradation
and high-resolution components. We show that extending this idea to
allow concatenation of features across up- and down-sampling stages 
(Dense DBPN) allows us to reconstruct further improve
super-resolution, yielding superior results and in particular
establishing new state of the art results for large scaling factors
such as $8\times$ across multiple data sets. 
\end{abstract}

\section{Introduction}
Significant progress in deep learning for
vision~\cite{huang2017densely,he2015deep,denton2015deep,shrivastava2016learning,larsson2016fractalnet,radford2015unsupervised,IMKDB17}
has recently been propagating to the field of super-resolution (SR)~\cite{johnson2016perceptual,liao2015video,dong2016image,Haris17,kappeler2016video,Kim_2016_VDSR,LapSRN,Tai-DRRN-2017}. 
\begin{figure}[t]
\centering
\includegraphics[width=8cm]{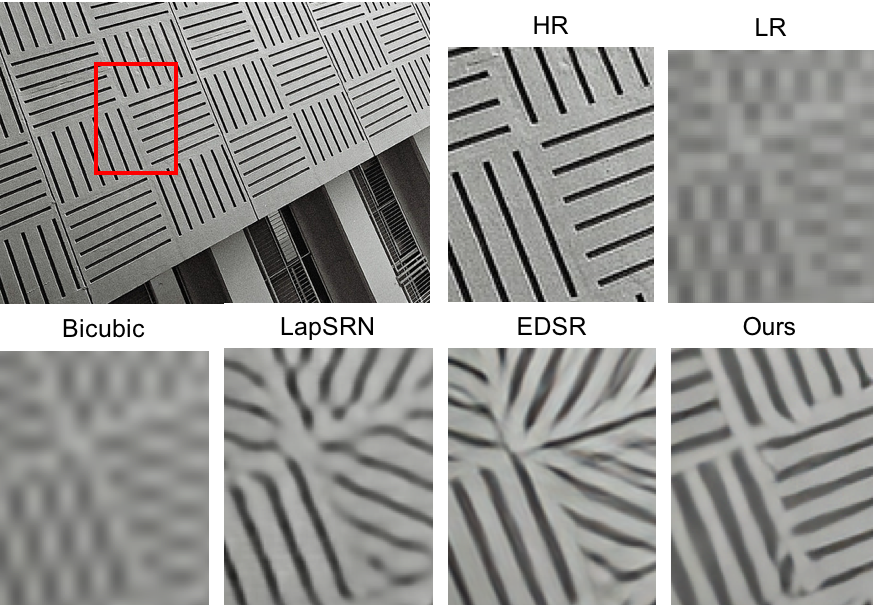}
\caption{Super-resolution result on $8\times$ enlargement. PSNR: LapSRN~\cite{LapSRN} (15.25 dB), EDSR~\cite{Lim_2017_CVPR_Workshops} (15.33 dB), and Ours (16.63 dB)}
\label{figure:intro}
\end{figure}
\begin{figure*}
\centering
\includegraphics[width=14.5cm]{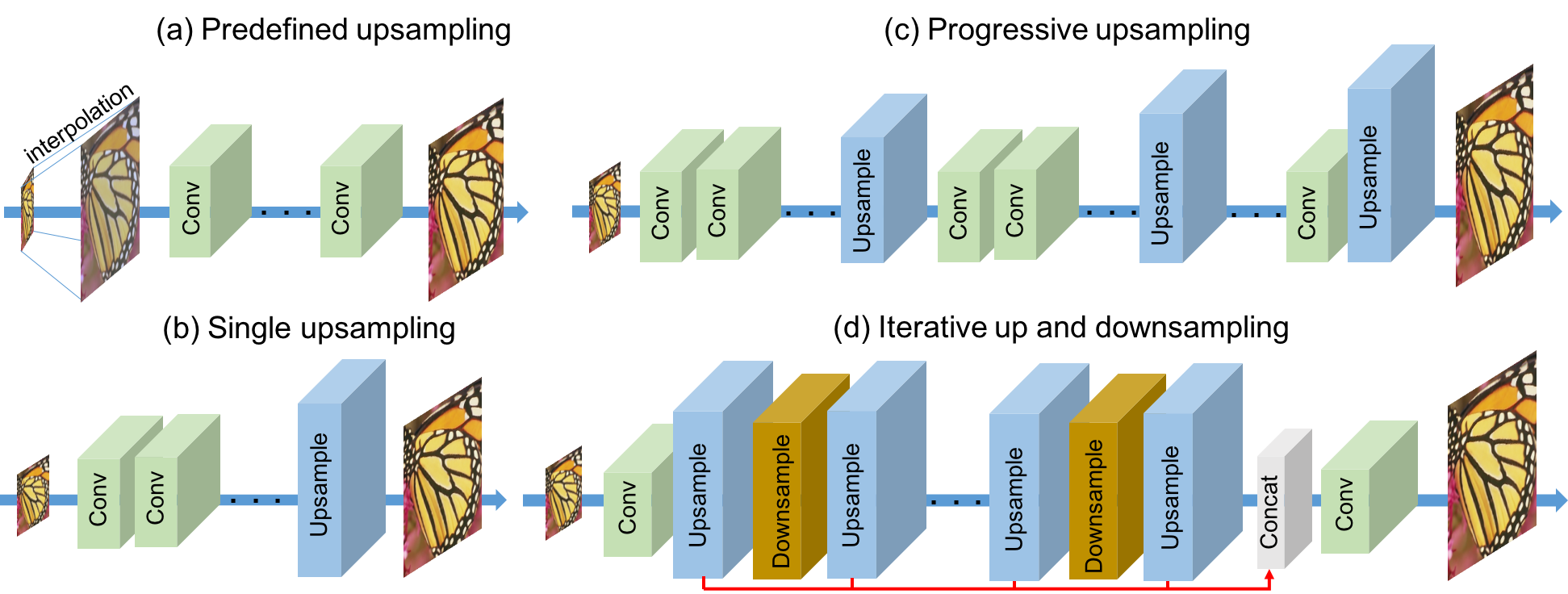}
\caption{Comparisons of Deep Network SR. (a) Predefined upsampling (e.g., SRCNN~\cite{dong2016image}, VDSR~\cite{Kim_2016_VDSR}, DRRN~\cite{Tai-DRRN-2017}) commonly uses the conventional interpolation, such as Bicubic, to upscale LR input images before entering the network. (b) Single upsampling (e.g., FSRCNN~\cite{dong2016accelerating}, ESPCN~\cite{shi2016real}) propagates the LR features, then construct the SR image at the last step. (c) Progressive upsampling uses a Laplacian pyramid network which gradually predicts SR images~\cite{LapSRN}. (d) Iterative up and downsampling approach is proposed by our DBPN which exploit the mutually connected up- (blue box) and down-sampling (gold box) stages to obtain numerous HR features in different depths.}
\label{figure:multi_upsampling}
\end{figure*}

Single image SR is an ill-posed inverse problem where the aim is to
recover a high-resolution (HR) image from a low-resolution (LR)
image. A currently typical approach is to construct an HR image by
learning non-linear LR-to-HR mapping, implemented as a deep neural
network~\cite{dong2016image, dong2016accelerating, shi2016real,
  LapSRN, Kim_2016_VDSR, kim2016deeply, Tai-DRRN-2017}. These networks
compute a sequence of feature maps from the LR image, culminating with
one or more upsampling layers to increase resolution and finally
construct the HR image. In contrast to this purely feed-forward
approach, human visual system is believed to use a feedback connection
to simply guide the task for the relevant
results~\cite{felleman1991distributed, kravitz2013ventral,
  lamme2000distinct}. Perhaps hampered by lack of such feedback,
the current SR networks with only feed-forward connections have difficulty in representing the LR to HR relation, especially for large scaling factors.

On the other hand, feedback connections were used effectively by one
of the early SR algorithms, the iterative
back-projection~\cite{irani1991improving}. It iteratively computes the
reconstruction error then fuses it back to tune the HR image
intensity. Although it has been proven to improve the image quality,
the result still suffers from ringing effect and chessboard
effect~\cite{dai2007bilateral}. Moreover, this method is sensitive to
choices of parameters such as the number of iterations and the blur
operator, leading to variability in results.




Inspired by~\cite{irani1991improving}, we construct an end-to-end
trainable architecture based on the idea of iterative up- and
down-sampling: Deep Back-Projection Networks (DBPN). Our networks
successfully perform large scaling factors, as shown
in~Fig.~\ref{figure:intro}. Our work provides
the following contributions:

\noindent(1) \textbf{Error feedback}. We propose an iterative
error-correcting feedback mechanism for SR, which calculates both up- and down-projection errors to guide the reconstruction for obtaining better results. Here, the projection errors are used to characterize or constraint the features in early layers. Detailed explanation can be seen in Section~\ref{sec:proposed}.

\noindent(2) \textbf{Mutually connected up- and down-sampling stages}. 
Feed-forward architectures, which is considered as a one-way mapping, only map rich representations of the input to the output space. This approach is unsuccessful to map LR and HR image, especially in large scaling factors, due to limited features available in the LR spaces. Therefore, our networks focus not only generating variants of the HR features using upsampling layers but also projecting it back to the LR spaces using downsampling layers. This connection is shown in~Fig.~\ref{figure:multi_upsampling} (d), alternating between up- (blue box) and down-sampling (gold box) stages, which represent the mutual relation of LR and HR image.

\noindent(3) \textbf{Deep concatenation}. Our networks represent different types of image degradation and HR components. This ability enables the networks to reconstruct the HR image using deep concatenation of the HR feature maps from all of the up-sampling steps. Unlike other networks, our reconstruction directly utilizes different types of LR-to-HR features without propagating them through the sampling layers as shown by the red arrow in~Fig.~\ref{figure:multi_upsampling} (d).

\noindent(4) \textbf{Improvement with dense connection}. We improve the accuracy of our network by densely connected~\cite{huang2017densely} each up- and down-sampling stage to encourage feature reuse.

\section{Related Work}
\subsection{Image super-resolution using deep networks}
Deep Networks SR can be primarily divided into four types as shown in~Fig.~\ref{figure:multi_upsampling}.

(a) \textbf{Predefined upsampling} commonly uses interpolation as the upsampling operator to produce middle resolution (MR) image. This schema was firstly proposed by SRCNN~\cite{dong2016image} to learn MR-to-HR non-linear mapping with simple convolutional layers. Later, the improved networks exploited residual learning~\cite{Kim_2016_VDSR,Tai-DRRN-2017} and recursive layers~\cite{kim2016deeply}. However, this approach might produce new noise from the MR image.

(b) \textbf{Single upsampling} offers simple yet effective way to increase the spatial resolution. This approach was proposed by FSRCNN~\cite{dong2016accelerating} and ESPCN~\cite{shi2016real}. 
These methods have been proven effective to increase the spatial resolution and replace predefined operators. However, they fail to learn complicated mapping due to limited capacity of the networks.
EDSR~\cite{Lim_2017_CVPR_Workshops}, the winner of NTIRE2017~\cite{timofte2017ntire}, belongs to this type. However, it
requires a large number of filters in each layer and lengthy training time, around eight days as stated by the authors. These problems open the opportunities to propose lighter networks that can preserve HR components better.

(c) \textbf{Progressive upsampling} was recently proposed in
LapSRN~\cite{LapSRN}. It progressively reconstructs the multiple SR
images with different scales in one feed-forward network. For the sake
of simplification, we can say that this network is the stacked of single upsampling networks which only relies on limited LR features. 
Due to this fact, LapSRN is outperformed even by our shallow networks especially for large scaling factors such as $8\times$ in experimental results.

(d) \textbf{Iterative up and downsampling} is proposed by our networks. We focus on increasing the sampling rate of SR features in different depths and distribute the tasks to calculate the reconstruction error to each stage. This schema enables the networks to preserve the HR components by learning various up- and down-sampling operators while generating deeper features.

\subsection{Feedback networks} 
Rather than learning a non-linear mapping of input-to-target space in one step, the feedback networks compose the prediction process into multiple steps which allow the model to have a self-correcting procedure. Feedback procedure has been implemented in various computing tasks~\cite{carreira2016human,ross2011learning,tu2010auto,li2016iterative,zamir2016feedback, shrivastava2016contextual,lotter2016deep}.

In the context of human pose estimation, Carreira et
al.~\cite{carreira2016human} proposed an iterative error feedback by
iteratively estimating and applying a correction to the current
estimation. PredNet~\cite{lotter2016deep} is an unsupervised recurrent
network to predictively code the future frames by recursively feeding
the predictions back into the model. For image segmentation, Li et
al.~\cite{li2016iterative} learn implicit shape priors and use them to
improve the prediction. However, to our knowledge, feedback procedures
have not been implemented to SR.

\subsection{Adversarial training}
Adversarial training, such as with Generative Adversarial Networks (GANs)~\cite{goodfellow2014generative} has been applied to various image reconstruction problems~\cite{ledig2016photo, sajjadi2016enhancenet, radford2015unsupervised, denton2015deep, johnson2016perceptual}. For the SR task, Johnson et al.~\cite{johnson2016perceptual} introduced perceptual losses based on high-level features extracted from pre-trained networks. Ledig et al.~\cite{ledig2016photo} proposed SRGAN which is considered as a single upsampling method. It proposed the natural image manifold that is able to create photo-realistic images by specifically formulating a loss function based on the euclidian distance between feature maps extracted from VGG19~\cite{simonyan2014very} and SRResNet. 

Our networks can be extended with the adversarial loss as generator network. However, we optimize our network only using an objective function such as mean square root error (MSE). Therefore, instead of training DBPN with the adversarial loss, we can compare DBPN with SRResNet which is also optimized by MSE.

\subsection{Back-projection} 
Back-projection~\cite{irani1991improving} is well known as the efficient iterative procedure to minimize the reconstruction error. Previous studies have proven the effectivity of back-projection~\cite{zhao2017iterative, haris2017first, dong2009nonlocal, timofte2016seven}. Originally, back-projection is designed for the case with multiple LR inputs. However, given only one LR input image, the updating procedure can be obtained by upsampling the LR image using multiple upsampling operators and calculate the reconstruction error iteratively~\cite{dai2007bilateral}. Timofte et al.~\cite{timofte2016seven} mentioned that back-projection can improve the quality of SR image. Zhao et al.~\cite{zhao2017iterative} proposed a method to refine high-frequency texture details with an iterative projection process. However, the initialization which leads to an optimal solution remains unknown. Most of the previous studies involve constant and unlearnable predefined parameters such as blur operator and number of iteration. 

To extend this algorithm, we develop an end-to-end trainable architecture which focuses to guide the SR task using mutually connected up- and down-sampling stages to learn non-linear relation of LR and HR image. The mutual relation between HR and LR image is constructed by creating iterative up and down-projection unit where the up-projection unit generates HR features, then the down-projection unit projects it back to the LR spaces as shown in~Fig.~\ref{figure:multi_upsampling} (d).
This schema enables the networks to preserve the HR components by learned various up- and down-sampling operators and generates deeper features to construct numerous LR and HR features.  

\section{Deep Back-Projection Networks} 
\label{sec:proposed}
Let $I^h$ and $I^l$ be HR and LR image with  $(M \times N)$ and $(
M^{'} \times N^{'})$, respectively, where $M^{'} < M$ and $N^{'} <
N$. The main building block of our proposed DBPN architecture is the
projection unit, which is trained (as part of the end-to-end training
of the SR system) to map either an LR feature map to an HR map
(up-projection), or an HR map to an LR map (down-projection). 

\subsection{Projection units}

The up-projection unit is defined as follows:
\begin{align}\label{eq:up-projection}
&\text{scale up:}&H^t_0 &= (L^{t-1} * p_{t})\uparrow_{s},\\
&\text{scale down:}&L^t_0 &= (H^t_0 * g_{t}) \downarrow_{s},\\
&\text{residual:}&e^l_t &= L^t_0 - L^{t-1},\\
&\text{scale residual up:}&H^t_1 &= (e^l_t* q_{t})\uparrow_{s},\\
&\text{output feature map:}&H^t &= H^t_0 + H^t_1&
\end{align}
where * is the spatial convolution operator, $\uparrow_{s}$ and $\downarrow_{s}$ are, respectively, the up- and
down-sampling operator with scaling factor $s$, and $p_t,
g_t, q_t$ are (de)convolutional layers at stage $t$.

\begin{figure}[t!]
\centering
\includegraphics[width=7.5cm]{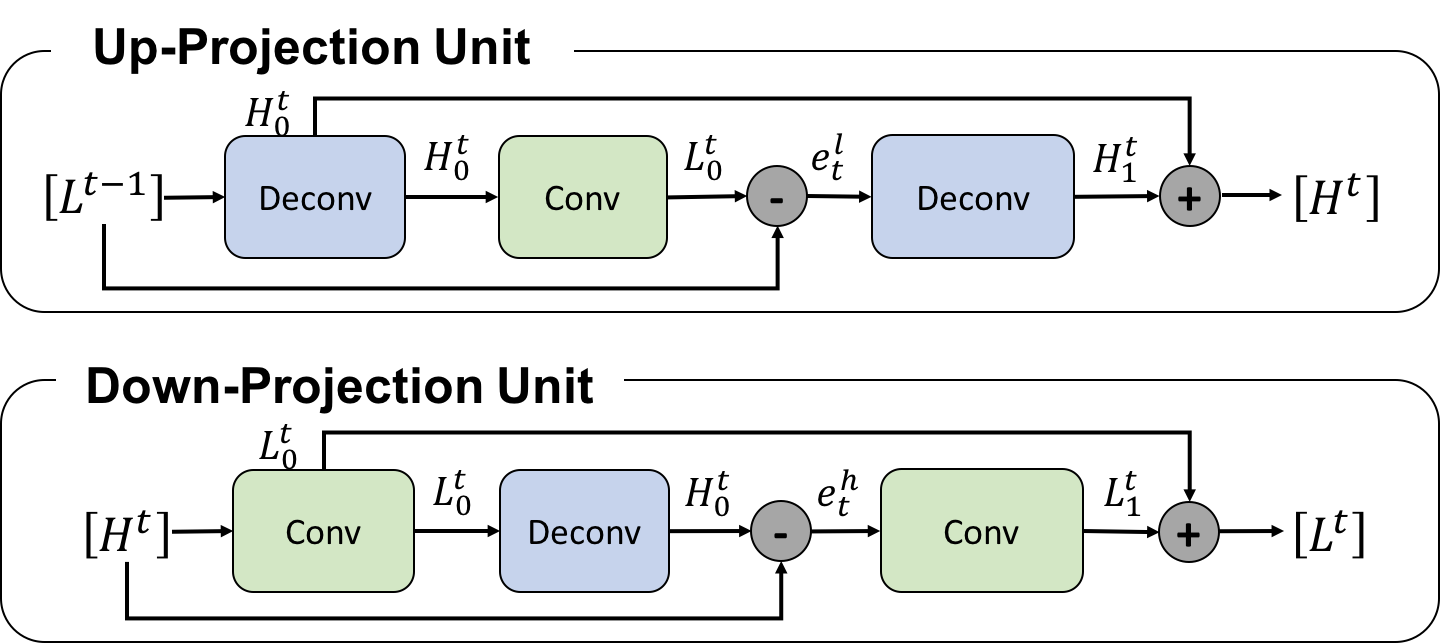}
\caption{Proposed up- and down-projection unit in the DBPN.}
\label{figure:projection_unit}
\end{figure}

This projection unit takes the previously computed LR feature map
$L^{t-1}$ as input, and maps it to an (intermediate) HR map $H^t_0$;
then it attempts to map it back to LR map $L^t_0$
(``back-project''). The residual (difference) $e^l_t$ between the observed LR map
$L^{t-1}$ and the reconstructed $L^t_0$ is mapped to HR again,
producing a new intermediate (residual) map $H^t_1$; the final output of the unit,
the HR map $H^t$, is obtained by summing the two intermediate HR maps. This step is illustrated in the upper part of~Fig.~\ref{figure:projection_unit}.

The down-projection unit is defined very similarly, but now its job is
to map its input HR map $H^t$ to the LR map $L^t$ as illustrated in the lower part of~Fig.~\ref{figure:projection_unit}.
\begin{align}\label{eq:down-projection}
&\text{scale down:}&L^t_0 &= (H^{t} * g'_{t})\downarrow_{s},\\
&\text{scale up:}&H^t_0 &= (L^t_0 * p'_{t}) \uparrow_{s},\\
&\text{residual:}&e^h_t &= H^t_0 - H^{t},\\
&\text{scale residual down:}&L^t_1 &= (e^h_t * g'_{t})\downarrow_{s},\\
&\text{output feature map:}&L^t &= L^t_0 + L^t_1\label{eq:proj-last}
\end{align}

We organize projection units in a series of \emph{stages}, alternating between $H$ and $L$.
These projection units can be understood as a self-correcting procedure which feeds a projection error to the sampling layer and iteratively changes the solution by feeding back the projection error.

The projection unit uses large sized filters such as $8\times 8$ and $12\times 12$. In other existing networks, the use of large-sized filter is avoided because it slows down the convergence speed and might produce sub-optimal results. However, iterative utilization of our projection units enables the network to suppress this limitation and to perform better performance on large scaling factor even with shallow networks.

\subsection{Dense projection units}
The dense inter-layer connectivity pattern in DenseNets~\cite{huang2017densely} has been shown to 
alleviate the vanishing-gradient problem, produce improved feature,
and encourage feature reuse. Inspired by this we propose to improve
DBPN, by introducing dense connections in the projection units called,
yielding Dense DBPN (D-DBPN).

Unlike the original DenseNets, we avoid dropout and batch norm, which are not suitable for SR, because they remove the range flexibility of the features~\cite{Lim_2017_CVPR_Workshops}.
Instead, we use $1 \times 1$ convolution layer as feature pooling and dimensional reduction~\cite{szegedy2015going,Haris17} before entering the projection unit.

\begin{figure}[t!]
\centering
\includegraphics[width=8.5cm]{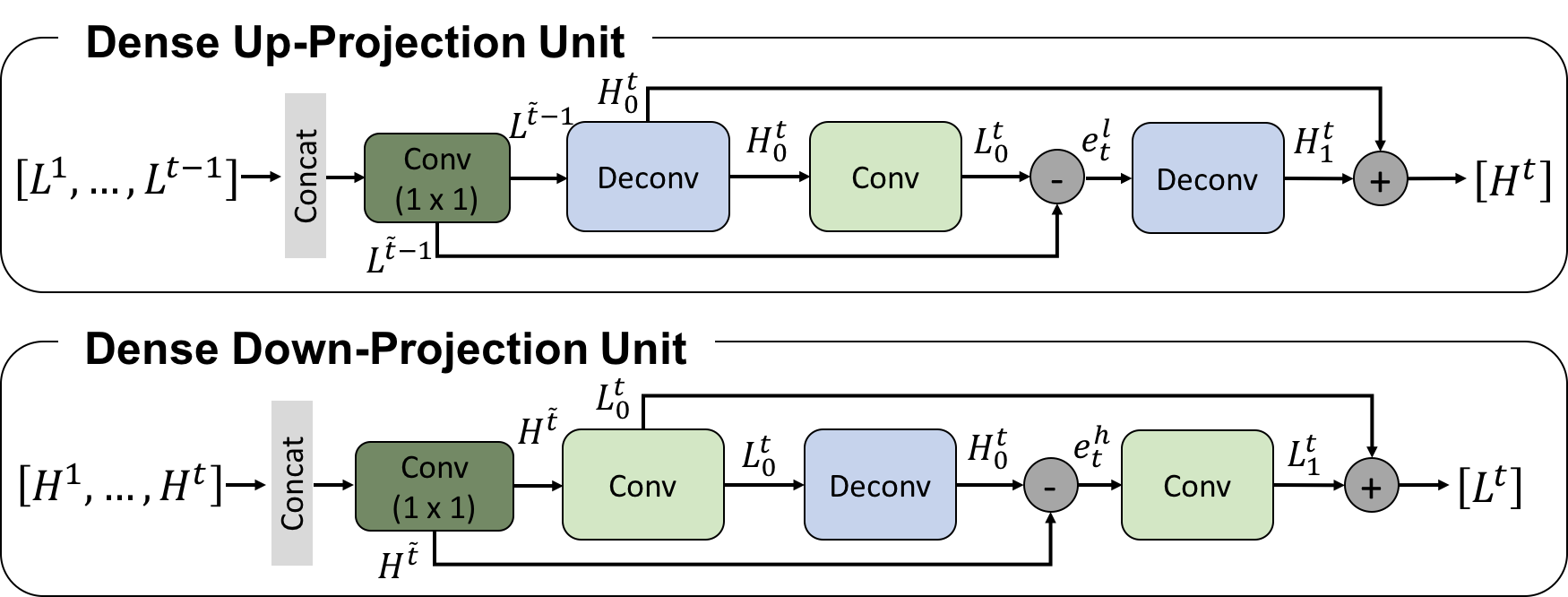}
\caption{Proposed up- and down-projection unit in the D-DBPN. The feature maps of all preceding units (i.e., $[L^{1}, ..., L^{t-1}]$ and $[H^{1}, ..., H^{t}]$ in up- and down-projections units, respectively) are concatenated and used as inputs, and its own feature maps are used as inputs into all subsequent units.}
\label{figure:D_DBPN}
\end{figure}

In D-DBPN, the input for each unit is the concatenation of the outputs from all previous units. Let the $L^{\tilde{t}}$ and $H^{\tilde{t}}$ be the input for dense up- and down-projection unit, respectively. They are generated using $conv(1,n_R)$ which is used to merge all previous outputs from each unit as shown in~Fig.~\ref{figure:D_DBPN}. This improvement enables us to generate the feature maps effectively, as shown in the experimental results.

\subsection{Network architecture}
\begin{figure*}[t]
\centering
\includegraphics[width=14cm]{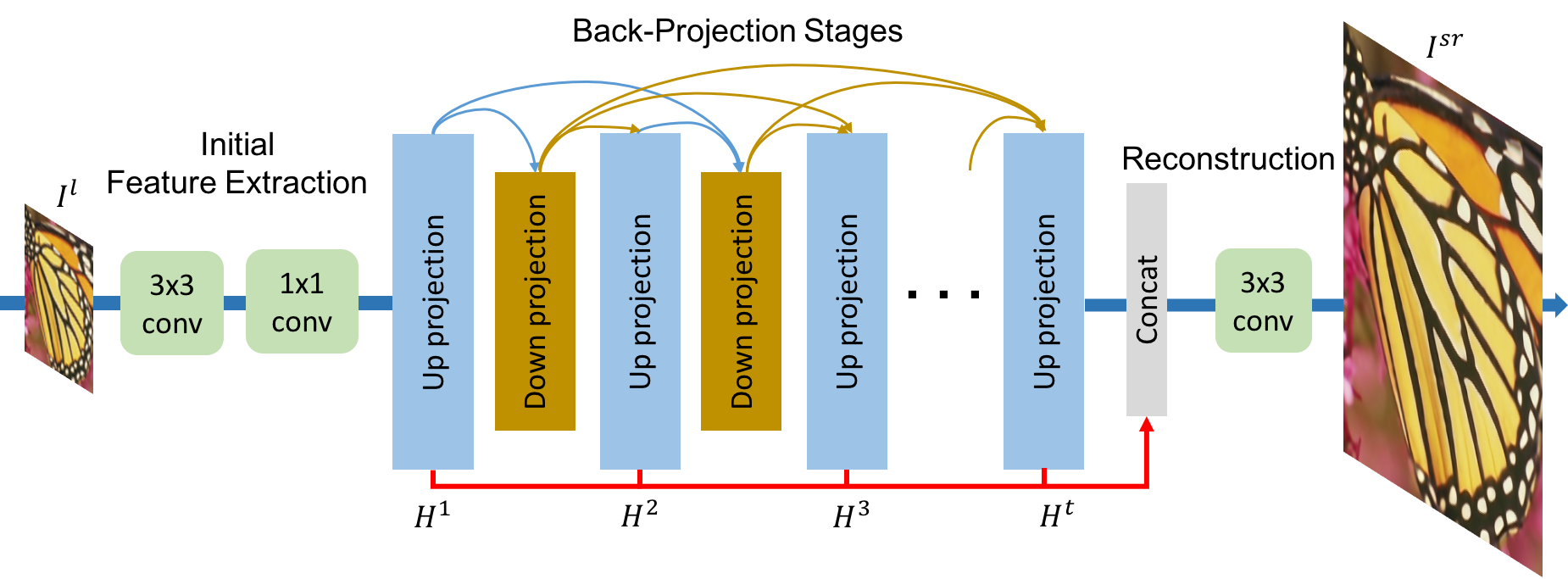}
\caption{An implementation of D-DBPN for super-resolution. Unlike the original DBPN, D-DBPN exploits densely connected projection unit to encourage feature reuse.}
\label{figure:proposed_network}
\end{figure*}

The proposed D-DBPN is illustrated in~Fig.~\ref{figure:proposed_network}. It can be divided into three
parts: initial feature extraction, projection, and reconstruction, as
described below. Here, let $conv(f,n )$ be a convolutional layer,
where $f$ is the filter size and $n$ is the number of filters. 

\begin{enumerate}
\item \textbf{Initial feature extraction}. We construct initial LR
  feature-maps $L^0$ from the input using $conv(3,n_{0})$. 
  Then $conv(1,n_R )$ is used to reduce the dimension from $n_{0}$ to $n_R$ before
  entering projection step where $n_0$ is the number of filters used in the initial LR features extraction and $n_R$ is the number of filters used in each projection unit.
\item \textbf{Back-projection stages}. Following initial feature
  extraction is a sequence of projection units, alternating between
  construction of LR and HR feature maps $H^t$,
  $L^t$; each unit has access to the outputs of all previous
  units.

\item \textbf{Reconstruction}. Finally, the target HR image is
  reconstructed as $I^{sr}=f_{Rec}([H^{1},H^{2},
  ..., H^{t}]),$ where $f_{Rec}$ use $conv(3,3)$ as
  reconstruction and $[H^{1},H^{2},
  ..., H^{t}]$ refers to the concatenation of the feature-maps
  produced in each up-projection unit.
\end{enumerate}

Due to the definitions of these building blocks, our network
architecture is modular. We can easily define and train networks with
different numbers of stages, controlling the depth. For a network with
$T$ stages, we have the initial extraction stage (2 layers), and then
$T$ up-projection units and $T-1$ down-projection units, each with 3
layers, followed by the reconstruction (one more layer). However, for the dense network, we add $conv(1,n_R)$ in each projection unit, except the first three units.

\section{Experimental Results}
\subsection{Implementation and training details}
In the proposed networks, the filter size in the projection unit is
various with respect to the scaling factor. For $2\times$ enlargement,
we use $6 \times 6$ convolutional layer with two striding and two
padding. Then, $4\times$ enlargement use $8 \times 8$ convolutional
layer with four striding and two padding. Finally, the $8\times$
enlargement use $12\times 12$ convolutional layer with eight striding
and two padding.\footnote{We found these settings to work well based
  on general intuition and preliminary experiments.}

We initialize the weights based on~\cite{he2015delving}. Here, std is computed by $(\sqrt{2/n_l})$ where $n_l=f^{2}_{t}n_{t}$, $f_{t}$ is the filter size, and $n_t$ is the number of filters. For example, with $f_{t}=3$ and $n_t=8$, the std is $0.111$. All convolutional and deconvolutional layers are followed by parametric rectified linear units (PReLUs).

We trained all networks using images from DIV2K \cite{timofte2017ntire}, Flickr \cite{Lim_2017_CVPR_Workshops}, and ImageNet dataset \cite{russakovsky2015imagenet} without augmentation.\footnote{The comparison on DIV2K only are available in the supplementary material.} To produce LR images, we downscale the HR images on particular scaling factors using Bicubic. We use batch size of 20 with size $32 \times 32$ for LR image, while HR image size corresponds to the scaling factors. The learning rate is initialized to $1e-4$ for all layers and decrease by a factor of 10 for every $5\times 10^5$ iterations for total $10^6$ iterations. For optimization, we use Adam with momentum to $0.9$ and weight decay to $1e-4$. All experiments were conducted using Caffe, MATLAB R2017a on NVIDIA TITAN X GPUs.

\subsection{Model analysis}
\label{subsec:modelanalysis}
\begin{figure}[t]
\centering
\includegraphics[width=8.5cm]{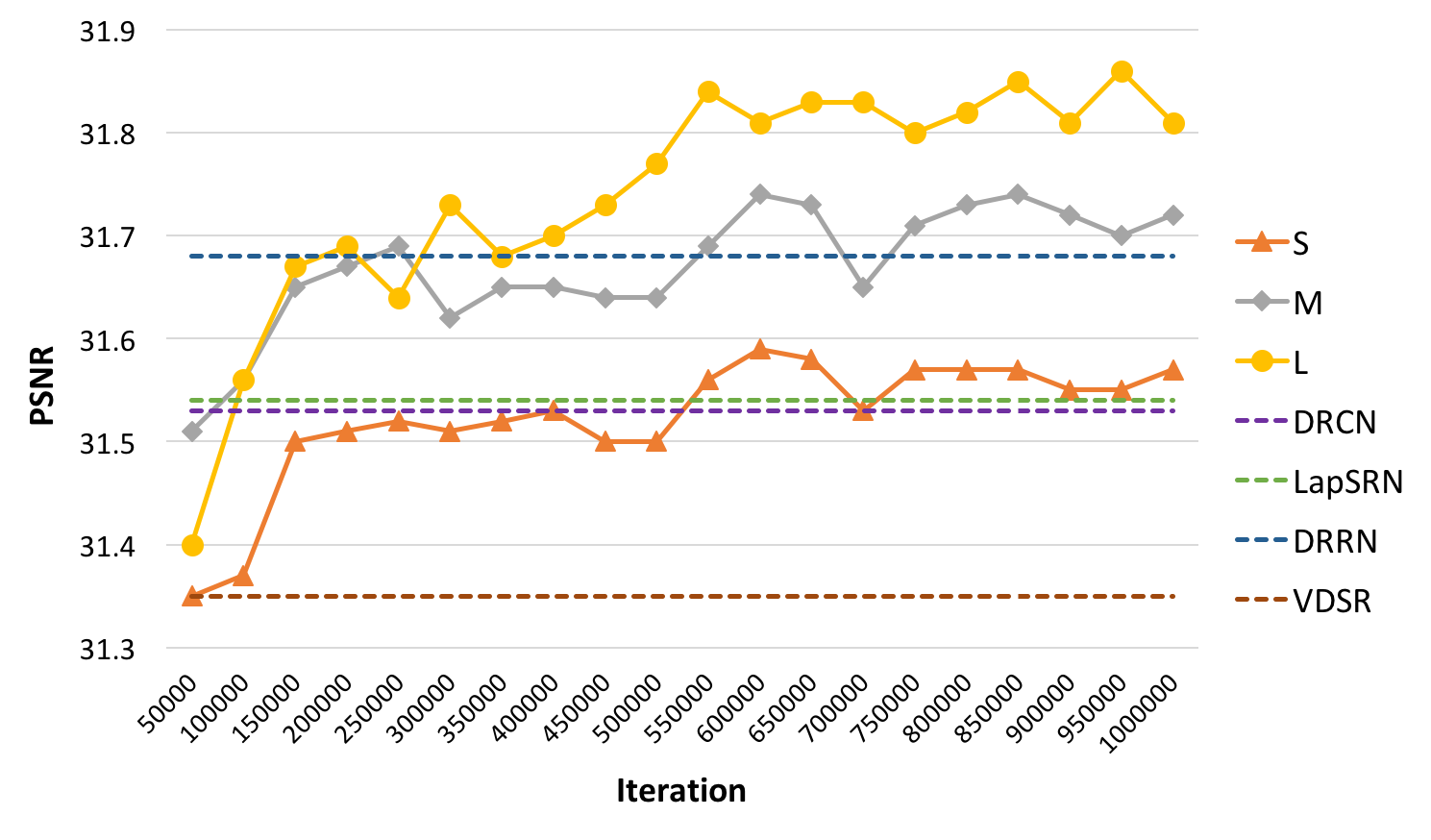}
\caption{The depth analysis of DBPNs compare to other networks (VDSR~\cite{Kim_2016_VDSR}, DRCN~\cite{kim2016deeply}, DRRN~\cite{Tai-DRRN-2017}, LapSRN~\cite{LapSRN}) on Set5 dataset for 4$\times$ enlargement.}
\label{figure:modular_comparison_4x}
\end{figure} 

\textbf{Depth analysis}. To demonstrate the capability of our projection unit, we construct multiple networks $S$ ($T=2$), $M$ ($T=4$), and $L$ ($T=6$) from the original DBPN. In the feature extraction, we use $conv(3,128)$ followed by $conv(1,32)$. Then, we use $conv(1,1)$ for the reconstruction. The input and output image are luminance only. 

The results on $4\times$ enlargement are shown in~Fig.~\ref{figure:modular_comparison_4x}. DBPN outperforms the state-of-the-art methods. Starting from our shallow network, the $S$ network gives the higher PSNR than VDSR, DRCN, and LapSRN. The $S$ network uses only 12 convolutional layers with smaller number of filters than VDSR, DRCN, and LapSRN. At the best performance, $S$ networks can achieve $31.59$ dB which better $0.24$ dB, $0.06$ dB, $0.05$ dB than VDSR, DRCN, and LapSRN, respectively. The $M$ network shows performance improvement which better than all four existing state-of-the-art methods (VDSR, DRCN, LapSRN, and DRRN). At the best performance, the $M$ network can achieve $31.74$ dB which better $0.39$ dB, $0.21$ dB, $0.20$ dB, $0.06$ dB than VDSR, DRCN, LapSRN, and DRRN respectively. In total, the $M$ network use 24 convolutional layers which has the same depth as LapSRN. Compare to DRRN (up to 52 convolutional layers), the $M$ network undeniable shows the effectiveness of our projection unit. Finally, the $L$ network outperforms all methods with $31.86$ dB which better $0.51$ dB, $0.33$ dB, $0.32$ dB, $0.18$ dB than VDSR, DRCN, LapSRN, and DRRN, respectively.

\begin{figure}[t]
\centering
\includegraphics[width=8.5cm]{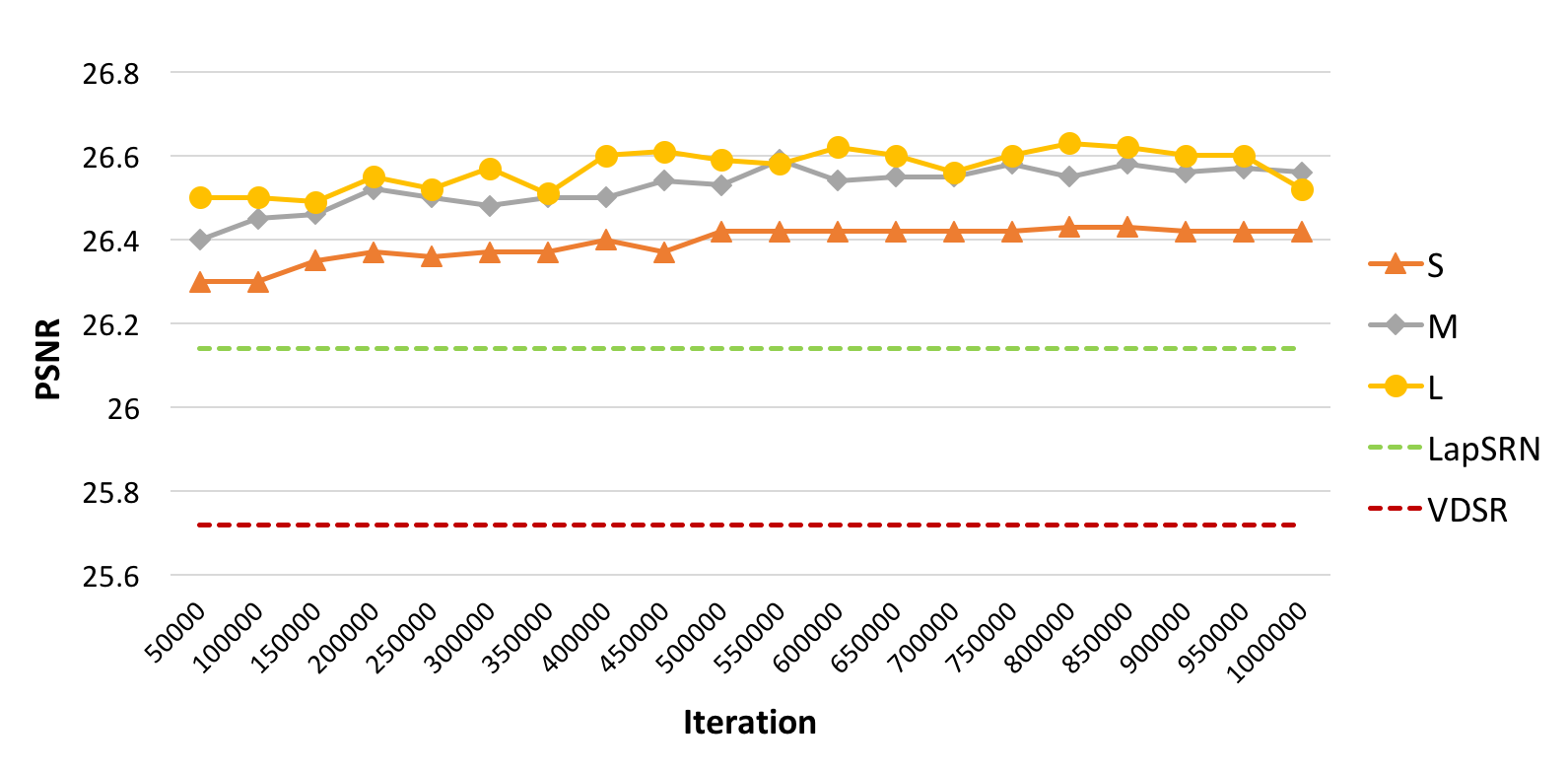}
\caption{The depth analysis of DBPN on Set5 dataset for 8$\times$ enlargement. S ($T=2$), M ($T=4$), and L ($T=6$)}
\label{figure:modular_comparison_8x}
\end{figure}

The results of $8\times$ enlargement are shown in~Fig.~\ref{figure:modular_comparison_8x}. The $S,M,L$ networks outperform the current state-of-the-art for $8\times$ enlargement which clearly show the effectiveness of our proposed networks on large scaling factors. However, we found that there is no significant performance gain from each proposed network especially for $L$ and $M$ networks where the difference only $0.04$ dB. 

\begin{figure}[t]
\centering
\includegraphics[width=8.5cm]{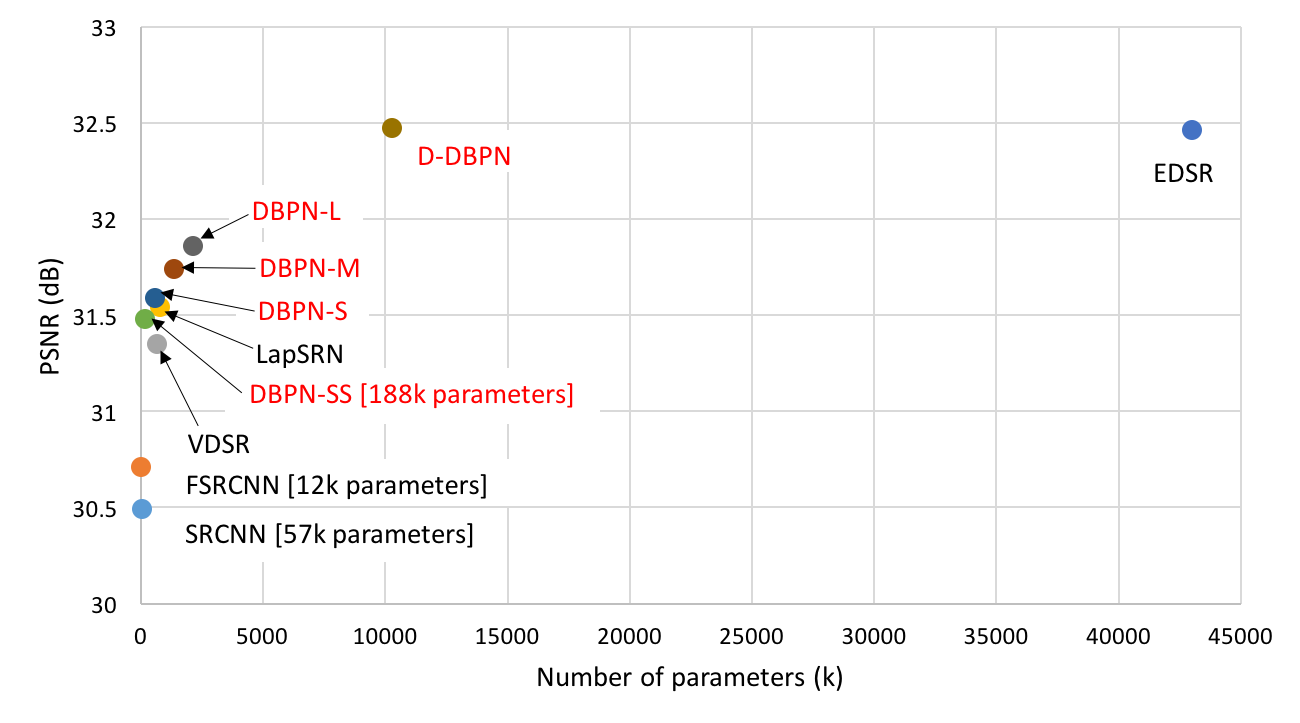}
\caption{Performance vs number of parameters. The results are evaluated with Set5 dataset for $4\times$ enlargement.}
\label{figure:psnr_vs_param_4x}

\includegraphics[width=8.5cm]{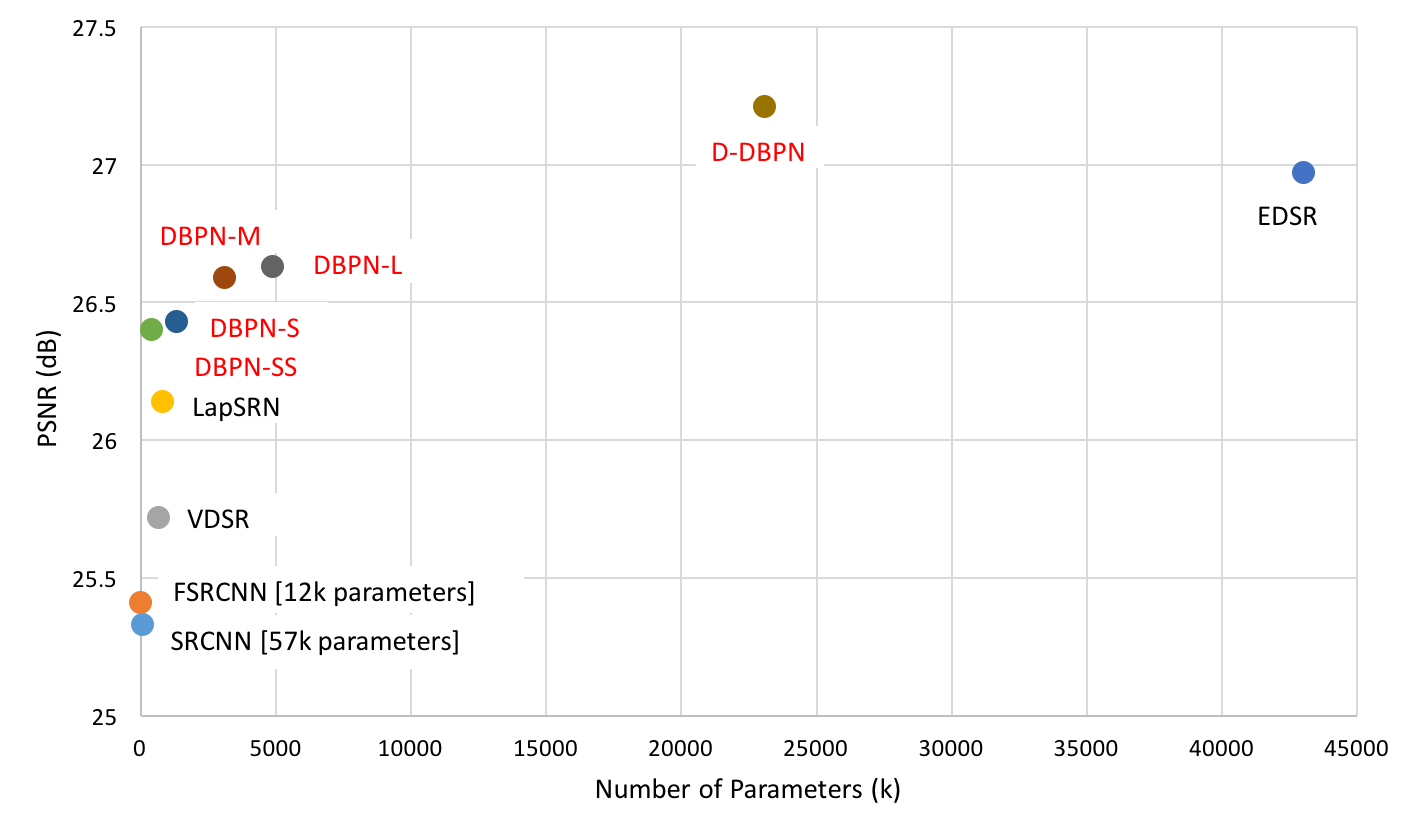}
\caption{Performance vs number of parameters. The results are evaluated with Set5 dataset for $8\times$ enlargement.}
\label{figure:psnr_vs_param_8x}
\end{figure}

\textbf{Number of parameters}. 
We show the tradeoff between performance
and number of network parameters from our networks and existing deep
network SR in~Fig.~\ref{figure:psnr_vs_param_4x}~and~\ref{figure:psnr_vs_param_8x}. 

For the sake of low computation for real-time processing, we construct $SS$ network which is the lighter version of the $S$ network, $(T=2)$. We only use $conv(3,64)$ followed by $conv(1,18)$ for the initial feature extraction. However, the results outperform SRCNN, FSRCNN, and VDSR on both $4\times$ and $8\times$ enlargement. Moreover, our $SS$ network performs better than VDSR with $72\%$ and $37\%$ fewer parameters on $4\times$ and $8\times$ enlargement, respectively. 

Our $S$ network has about $27\%$ fewer parameters and higher PSNR than LapSRN on $4\times$ enlargement. 
Finally, D-DBPN has about $76\%$ fewer parameters, and
approximately the same PSNR, compared to EDSR on $4\times$ enlargement. On the $8\times$ enlargement, D-DBPN has about $47\%$ fewer parameters with better PSNR compare to EDSR. This evidence show that our networks has the best trade-off between performance and number of parameter.

\begin{figure}[t]
\centering
\includegraphics[width=8.5cm]{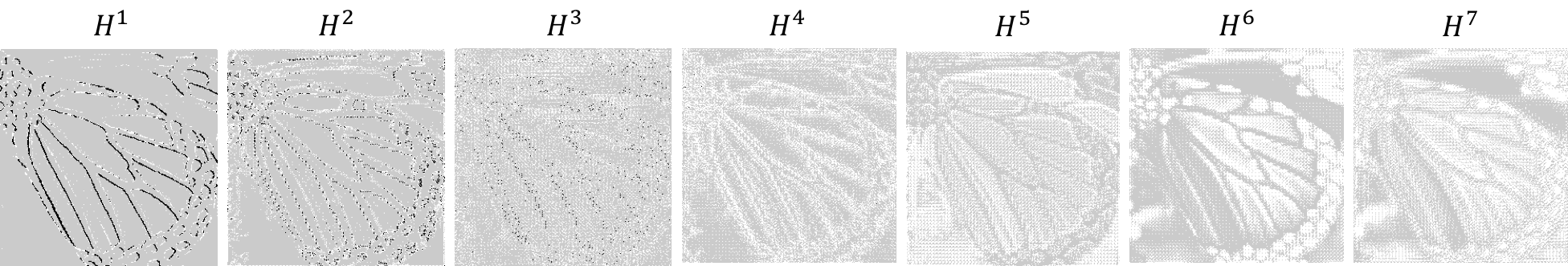}
\caption{Sample of activation maps from up-projection units in D-DBPN where $t=7$. Each feature has been enhanced using the same grayscale colormap for visibility.}
\label{figure:result_up_projection}
\end{figure}

\textbf{Deep concatenation}. Each projection unit is used to
distribute the reconstruction step by constructing features
which represent different details of the HR components. Deep concatenation is also well-related with the number of $T$ (back-projection stage), 
which shows more detailed features generated from the projection units will also increase the quality of the results. In~Fig.~\ref{figure:result_up_projection}, it is shown that each stage successfully generates diverse features to reconstruct SR image.

\textbf{Dense connection}. We implement D-DBPN-L which is a dense connection of the $L$ network to show how dense connection can improve the network's performance in all cases as shown in~Table~\ref{tab:dense}. On $4\times$ enlargement, the dense network, D-DBPN-L, gains $0.13$ dB and $0.05$ dB higher than DBPN-L on the Set5 and Set14, respectively. On $8\times$, the gaps are even larger. The D-DBPN-L has $0.23$ dB and $0.19$ dB higher that DBPN-L on the Set5 and Set14, respectively.
\begin{table}[h!]
\small
\caption{Comparison of the DBPN-L and D-DBPN-L on 4$\times$ and 8$\times$ enlargement. {\color{red}Red} indicates the best performance.}
\centering
\label{tab:dense}
\begin{tabular}{*1l*1c|*2c*2c}
\hline\noalign{\smallskip}
\smallskip & &\multicolumn{2}{c}{Set5} & \multicolumn{2}{c}{Set14} \\         
Algorithm & Scale & PSNR&SSIM & PSNR&SSIM   \\
\noalign{\smallskip}\hline\noalign{\smallskip}
DBPN-L&4		&$31.86$&$0.891$&$28.47$&$0.777$\\
D-DBPN-L&4		&{\color{red}$31.99$}&{\color{red}$0.893$}&{\color{red}$28.52$}&{\color{red}$0.778$}\\
\noalign{\smallskip}\hline\noalign{\smallskip}
DBPN-L&8		&$26.63$&$0.761$&$24.73$&$0.631$\\
D-DBPN-L&8		&{\color{red}$26.86$}&{\color{red}$0.773$}&{\color{red}$24.92$}&{\color{red}$0.638$}\\
\noalign{\smallskip}\hline
\end{tabular}
\end{table}


\subsection{Comparison with the-state-of-the-arts}
\begin{figure*}
\centering
\includegraphics[width=15cm]{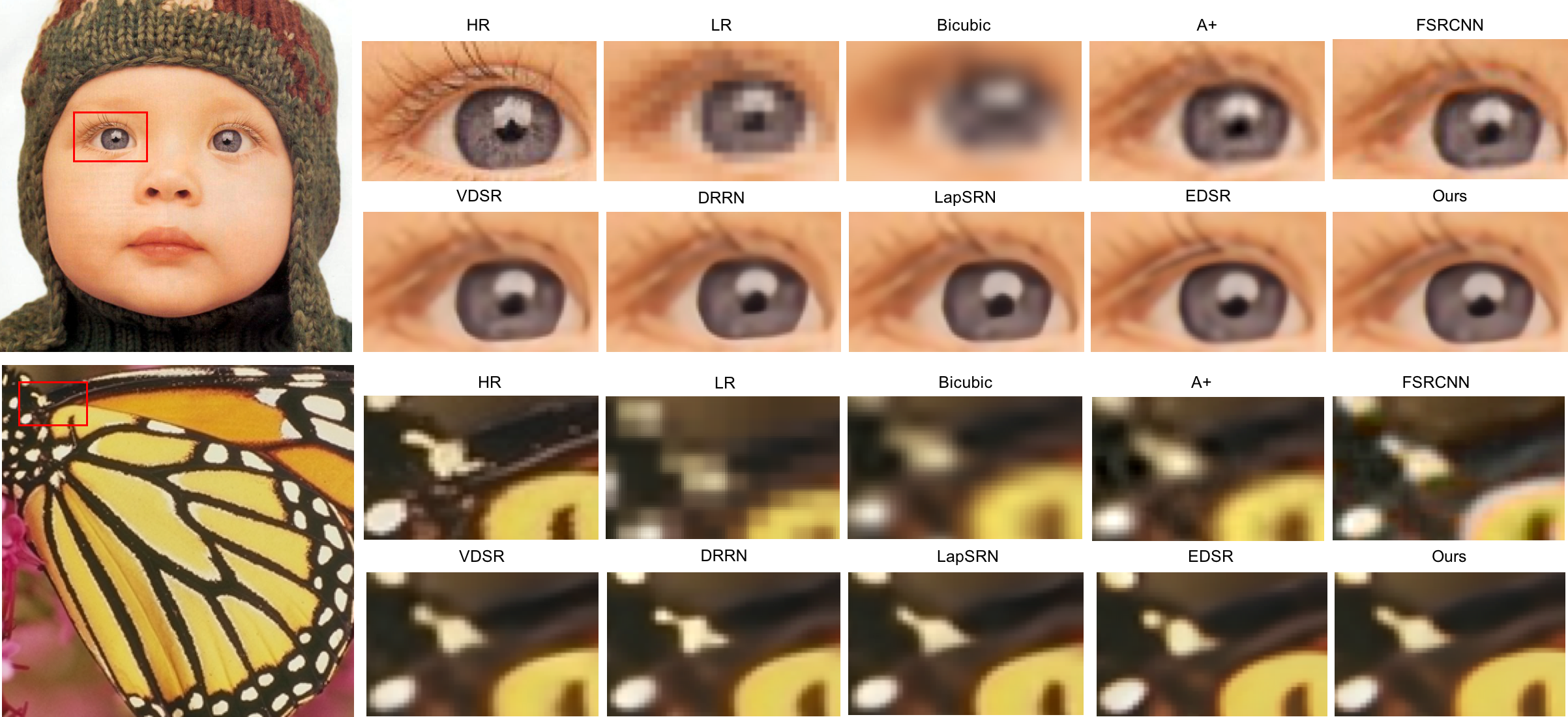}
\caption{Qualitative comparison of our models with other works on $4\times$ super-resolution.}
\label{figure:4x_result}
\end{figure*}

To confirm the ability of the proposed network, we performed several experiments and analysis. We compare our network with eight state-of-the-art SR algorithms: A+~\cite{timofte2014a+}, SRCNN~\cite{dong2016image}, FSRCNN~\cite{dong2016accelerating}, VDSR~\cite{Kim_2016_VDSR}, DRCN~\cite{kim2016deeply}, DRRN~\cite{Tai-DRRN-2017}, LapSRN~\cite{LapSRN}, and EDSR~\cite{Lim_2017_CVPR_Workshops}. We carry out extensive experiments using 5 datasets: Set5~\cite{bevilacqua2012low}, Set14~\cite{zeyde2012single}, BSDS100~\cite{arbelaez2011contour}, Urban100~\cite{huang2015single} and Manga109~\cite{matsui2016sketch}. Each dataset has different characteristics. Set5, Set14 and BSDS100 consist of natural scenes; Urban100 contains urban scenes with details in different frequency bands; and Manga109 is a dataset of Japanese manga. Due to computation limit of Caffe, we have to divide each image in Urban100 and Manga109 into four parts and then calculate PSNR separately.

Our final network, D-DBPN, uses $conv(3,256)$ then $conv(1,64)$ for the initial feature extraction and $t=7$ for the back-projection stages. In the reconstruction, we use $conv(3,3)$. RGB color channels are used for input and output image. It takes less than four days to train.

PSNR~\cite{irani93} and structural similarity (SSIM)~\cite{wang04}
were used to quantitatively evaluate the proposed method. Note that higher PSNR
and SSIM values indicate better quality. As used by existing networks,
all measurements used only the luminance channel (Y). For SR by factor
$s$, we crop $s$ pixels near image boundary before evaluation as in~\cite{Lim_2017_CVPR_Workshops, dong2016accelerating}. Some of the existing networks such as SRCNN, FSRCNN, VDSR, and EDSR did not perform $8\times$ enlargement. To this end, we retrained the existing networks by using author's code with the recommended parameters. 

We show the quantitative results in the~Table~\ref{tab:psnr}. Our D-DBPN outperforms the existing methods by a large margin in all scales except EDSR. For the $2\times$ and $4\times$ enlargement, we have comparable PSNR with EDSR. However, the result of EDSR tends to generate stronger edge than the ground truth and lead to misleading information in several cases. The result of EDSR for eyelashes in~Fig.~\ref{figure:4x_result} shows that it was interpreted as a stripe pattern. On the other hand, our result generates softer patterns which subjectively closer to the ground truth. On the butterfly image, EDSR separates the white pattern which shows that EDSR tends to construct regular pattern such ac circle and stripe, while D-DBPN constructs the same pattern as the ground truth. The previous statement is strengthened by the results from the Urban100 dataset which consist of many regular patterns from buildings. In Urban100, EDSR has $0.54$ dB higher than D-DBPN.

Our network shows it's effectiveness in the $8\times$ enlargement. The D-DBPN outperforms all of the existing methods by a large margin. Interesting results are shown on Manga109 dataset where D-DBPN obtains $25.50$ dB which is $0.61$ dB better than EDSR. While on the Urban100 dataset, D-DBPN achieves 23.25 which is only $0.13$ dB better than EDSR. The results show that our networks perform better on fine-structures images such as manga characters, even though we do not use any animation images in the training.

\begin{figure*}
\centering
\includegraphics[width=14.5cm]{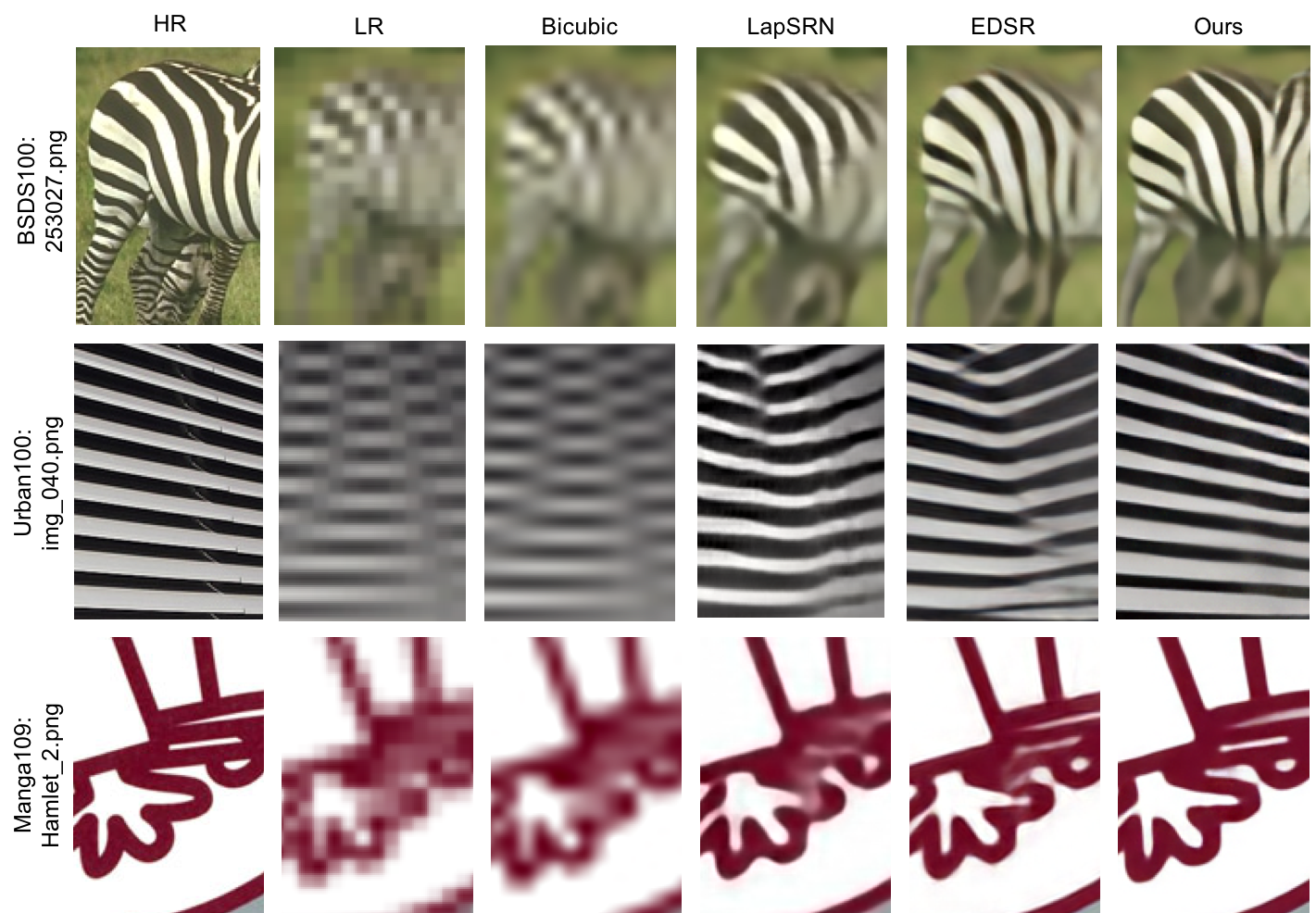}
\caption{Qualitative comparison of our models with other works on $8\times$ super-resolution. $1^{st}$ line: LapSRN~\cite{LapSRN} (19.77 dB), EDSR~\cite{Lim_2017_CVPR_Workshops} (19.79 dB), and Ours (19.82 dB). $2^{nd}$ line: LapSRN~\cite{LapSRN} (16.45 dB), EDSR~\cite{Lim_2017_CVPR_Workshops} (19.1 dB), and Ours (23.1 dB). $3^{rd}$ line: LapSRN~\cite{LapSRN} (24.34 dB), EDSR~\cite{Lim_2017_CVPR_Workshops} (25.29 dB), and Ours (28.84 dB)}
\label{figure:8x_result}
\end{figure*}

The results of $8\times$ enlargement are visually shown in~Fig.~\ref{figure:8x_result}. Qualitatively, D-DBPN is able to preserve the HR components better than other networks. It shows that our networks can extract not only features but also create contextual information from the LR input to generate HR components in the case of large scaling factors, such as $8\times$ enlargement.


\begin{table*}[t!]
\scriptsize
\caption{Quantitative evaluation of state-of-the-art SR algorithms: average PSNR/SSIM for scale factors 2$\times$, 4$\times$ and 8$\times$. {\color{red}Red} indicates the best and {\color{blue}blue} indicates the second best performance. (* indicates that the input is divided into four parts and calculated separately due to computation limitation of Caffe)}
\centering
\label{tab:psnr}
\begin{tabular}{*1l*1c*2c*2c*2c*2c*2c}
\hline\noalign{\smallskip}
\smallskip & &\multicolumn{2}{c}{Set5} & \multicolumn{2}{c}{Set14}& \multicolumn{2}{c}{BSDS100}& \multicolumn{2}{c}{Urban100}&\multicolumn{2}{c}{Manga109} \\         
Algorithm & Scale & PSNR&SSIM & PSNR&SSIM & PSNR&SSIM & PSNR&SSIM & PSNR&SSIM  \\
\noalign{\smallskip}\hline\noalign{\smallskip}
Bicubic&2								&$33.65$&$0.930$&$30.34$&$0.870$&$29.56$&$0.844$&$26.88 \hspace{0.5mm}(27.39^*)$&$0.841$&$30.84 \hspace{0.5mm}(31.05^*)$&$0.935$\\
A+~\cite{timofte2014a+}&2					&$36.54$&$0.954$&$32.40$&$0.906$&$31.22$&$0.887$&$29.23$&$0.894$&$35.33$&$0.967$\\
SRCNN~\cite{dong2016image}&2			&$36.65$&$0.954$&$32.29$&$0.903$&$31.36$&$0.888$&$29.52$&$0.895$&$35.72$&$0.968$\\
FSRCNN~\cite{dong2016accelerating}&2		&$36.99$&$0.955$&$32.73$&$0.909$&$31.51$&$0.891$&$29.87$&$0.901$&$36.62$&$0.971$\\
VDSR~\cite{Kim_2016_VDSR}&2			&$37.53$&$0.958$&$32.97$&$0.913$&$31.90$&$0.896$&$30.77$&$0.914$&$37.16$&$0.974$\\
DRCN~\cite{kim2016deeply}&2				&$37.63$&$0.959$&$32.98$&$0.913$&$31.85$&$0.894$&$30.76$&$0.913$&$37.57$&$0.973$\\
DRRN~\cite{Tai-DRRN-2017}&2				&$37.74$&$0.959$&$33.23$&$0.913$&$32.05$&$0.897$&$31.23$&$0.919$&$37.92$&$0.976$\\
LapSRN~\cite{LapSRN}&2					&$37.52$&$0.959$&$33.08$&$0.913$&$31.80$&$0.895$&$30.41\hspace{0.5mm} (31.05^*)$&$0.910$&$37.27 \hspace{0.5mm}(37.53^*)$&$0.974$\\
EDSR~\cite{Lim_2017_CVPR_Workshops}&2	&{\color{red}$38.11$}&{\color{red}$0.960$}&{\color{red}$33.92$}&{\color{red}$0.919$}&{\color{red}$32.32$}&{\color{red}$0.901$}&{\color{red}$32.93\hspace{0.5mm}(33.56^*)$}&{\color{red}$0.935$}&{\color{red}$39.10\hspace{0.5mm}(39.33^*)$}&{\color{blue}$0.977$}\\
D-DBPN &2							&{\color{blue}$38.09$}&{\color{red}$0.960$}&{\color{blue}$33.85$}&{\color{red}$0.919$}&{\color{blue}$32.27$}&{\color{blue}$0.900$}&{\color{blue}$\hspace{2.5mm}-\hspace{2.5mm}(33.02^*)$}&{\color{blue}$0.931$}&{\color{blue}$\hspace{2.5mm}-\hspace{2.5mm}(39.32^*)$}&{\color{red}$0.978$}\\
\noalign{\smallskip}\hline\noalign{\smallskip}
Bicubic&4								&$28.42$&$0.810$&$26.10$&$0.704$&$25.96$&$0.669$&$23.15\hspace{0.5mm} (23.64^*)$&$0.659$&$24.92\hspace{0.5mm} (25.15^*)$&$0.789$\\
A+~\cite{timofte2014a+}&4					&$30.30$&$0.859$&$27.43$&$0.752$&$26.82$&$0.710$&$24.34$&$0.720$&$27.02$&$0.850$\\
SRCNN~\cite{dong2016image}&4			&$30.49$&$0.862$&$27.61$&$0.754$&$26.91$&$0.712$&$24.53$&$0.724$&$27.66$&$0.858$\\
FSRCNN~\cite{dong2016accelerating}&4		&$30.71$&$0.865$&$27.70$&$0.756$&$26.97$&$0.714$&$24.61$&$0.727$&$27.89$&$0.859$\\
VDSR~\cite{Kim_2016_VDSR}&4			&$31.35$&$0.882$&$28.03$&$0.770$&$27.29$&$0.726$&$25.18$&$0.753$&$28.82$&$0.886$\\
DRCN~\cite{kim2016deeply}&4				&$31.53$&$0.884$&$28.04$&$0.770$&$27.24$&$0.724$&$25.14$&$0.752$&$28.97$&$0.886$\\
DRRN~\cite{Tai-DRRN-2017}&4				&$31.68$&$0.888$&$28.21$&$0.772$&$27.38$&$0.728$&$25.44$&$0.764$&$29.46$&$0.896$\\
LapSRN~\cite{LapSRN}&4					&$31.54$&$0.885$&$28.19$&$0.772$&$27.32$&$0.728$&$25.21\hspace{0.5mm} (25.87^*)$&$0.756$&$29.09\hspace{0.5mm} (29.44^*)$&$0.890$\\
EDSR~\cite{Lim_2017_CVPR_Workshops}&4	&{\color{blue}$32.46$}&{\color{blue}$0.897$}&{\color{blue}$28.80$}&{\color{red}$0.788$}&{\color{blue}$27.71$}&{\color{red}$0.742$}&{\color{red}$26.64\hspace{0.5mm}(27.30^*)$}&{\color{red}$0.803$}&{\color{blue}$31.02\hspace{0.5mm}(31.41^*)$}&{\color{red}$0.915$}\\
D-DBPN &4							&{\color{red}$32.47$}&{\color{red}$0.898$}&{\color{red}$28.82$}&{\color{blue}$0.786$}&{\color{red}$27.72$}&{\color{blue}$0.740$}&{\color{blue}$\hspace{2.5mm}-\hspace{2.5mm}(27.08^*)$}&{\color{blue}$0.795$}&{\color{red}$\hspace{2.5mm}-\hspace{2.5mm}(31.50^*)$}&{\color{blue}$0.914$}\\
\noalign{\smallskip}\hline\noalign{\smallskip}
Bicubic&8								&$24.39$&$0.657$&$23.19$&$0.568$&$23.67$&$0.547$&$20.74\hspace{0.5mm} (21.24^*)$&$0.516$&$21.47\hspace{0.5mm} (21.68^*)$&$0.647$\\
A+~\cite{timofte2014a+}&8					&$25.52$&$0.692$&$23.98$&$0.597$&$24.20$&$0.568$&$21.37$&$0.545$&$22.39$&$0.680$\\
SRCNN~\cite{dong2016image}&8			&$25.33$&$0.689$&$23.85$&$0.593$&$24.13$&$0.565$&$21.29$&$0.543$&$22.37$&$0.682$\\
FSRCNN~\cite{dong2016accelerating}&8		&$25.41$&$0.682$&$23.93$&$0.592$&$24.21$&$0.567$&$21.32$&$0.537$&$22.39$&$0.672$\\
VDSR~\cite{Kim_2016_VDSR}&8			&$25.72$&$0.711$&$24.21$&$0.609$&$24.37$&$0.576$&$21.54$&$0.560$&$22.83$&$0.707$\\
LapSRN~\cite{LapSRN}&8					&$26.14$&$0.738$&$24.44$&$0.623$&$24.54$&$0.586$&$21.81\hspace{0.5mm} (22.42^*)$&$0.582$&$23.39 \hspace{0.5mm}(23.67^*)$&$0.735$\\
EDSR~\cite{Lim_2017_CVPR_Workshops}&8	&{\color{blue}$26.97$}&{\color{blue}$0.775$}&{\color{blue}$24.94$}&{\color{blue}$0.640$}&{\color{blue}$24.80$}&{\color{blue}$0.596$}&{\color{blue}$22.47\hspace{0.5mm}(23.12^*)$}&{\color{blue}$0.620$}&{\color{blue}$ 24.58\hspace{0.5mm}(24.89^*)$}&{\color{blue}$0.778$}\\
D-DBPN &8							&{\color{red}$27.21$}&{\color{red}$0.784$}&{\color{red}$25.13$}&{\color{red}$0.648$}&{\color{red}$24.88$}&{\color{red}$0.601$}&{\color{red}$\hspace{2.5mm}-\hspace{2.5mm}(23.25^*)$}&{\color{red}$0.622$}&{\color{red}$\hspace{2.5mm}-\hspace{2.5mm}(25.50^*)$}&{\color{red}$0.799$}\\
\noalign{\smallskip}\hline
\end{tabular}
\end{table*}

\section{Conclusion}
We have proposed Deep Back-Projection Networks for Single Image Super-resolution. Unlike the previous methods which predict the SR image in a feed-forward manner, our proposed networks focus to directly increase the SR features using multiple up- and down-sampling stages and feed the error predictions on each depth in the networks to revise the sampling results, then, accumulates the self-correcting features from each upsampling stage to create SR image. We use error feedbacks from the up- and down-scaling steps to guide the network to achieve a better result. The results show the effectiveness of the proposed network compares to other state-of-the-art methods.  Moreover, our proposed network successfully outperforms other state-of-the-art methods on large scaling factors such as $8\times$ enlargement. 


{\small
\bibliographystyle{ieee}
\bibliography{egbib}
}

\end{document}